# Graph Coloring Using Heat Diffusion

Vivek Chaudhary [0000-0002-5517-3190]

vivekch2018@gmail.com

**Abstract.** Graph coloring is a problem with varied applications in industry and science such as scheduling, resource allocation, and circuit design. The purpose of this paper is to establish if a new gradient based iterative solver framework known as heat diffusion can solve the graph coloring problem. We propose a solution to the graph coloring problem using the heat diffusion framework. We compare the solutions against popular methods and establish the competitiveness of heat diffusion method for the graph coloring problem.

**Keywords:** Combinatorial Optimization, Graph Coloring, Machine Learning

## 1 Introduction

Combinatorial optimization finds applications in diverse fields such as supply chain [1], traffic flow optimization [2], molecular dynamic analysis [3], and financial risk analysis [4]. With the explosion in the machine learning research in recent decades, there has been marked interest in use of machine learning for solving combinatorial optimization problems [5]. Machine learning approaches for tackling combinatorial optimization can be broadly classified into three classes. First is supervised learning. Supervised learning is arguably the more popular among the three classes [6][7][8]. However, supervised learning requires training sets of solved instances. Getting large training sets would involve solving large number of problems, which is extremely resource intensive [9]. The second class of machine learning approaches is reinforcement learning. Reinforcement learning has shown some very exciting results despite operating in discrete action spaces, mastering video games [10], and other games such as chess and go [11]. However, reinforcement learning approaches lack full differentiability [18], hence making learning resource intensive. The third machine learning approach is unsupervised learning. The challenge to unsupervised learning is that it requires bespoke loss functions to guide learning. There have been several creative attempts at solving this problem [12]. A new approach for solving combinatorial optimizations was proposed in [13], where the unsupervised learning is framed as an iterative approximation solver. This approach has achieved high efficiency by propagating information from the scope of search effectively using heat diffusion.

This paper builds on [13] to solve the graph coloring problem [14], one of the most challenging problems in combinatorial optimization [15]. Graph coloring finds appli-cations in areas such as scheduling [16] and register allocation [17]. The details of the graph coloring problem are provided in section 2. The heat diffusion framework is detailed in section 3. Section 4 contains details of our experiments along with the results. We show that our method is competitive with popular methods. The paper is concluded in section 5. The codebase for reproducing all experiment results is available at https://github.com/chaudhary-vivek/HeO_GCP.

## 2       Graph Coloring Problem

### 2.1     Problem definition

Considering an undirected graph $G = (V,E)$, where $V = \{1,2,3,......,n\}$ is the set of vertices and $E = \{(i,j) : i, j \in V\}$ is the set of edges. In the graph coloring problem, we assign an integer $c(v) \in \{1,2,3,.....q\}$ to each vertex $v \in V$, such that no two adjacent vertices are assigned the same integer, $c(i) \neq c(j)\ \forall (i,j) \in E,$ while using at most $q$ integers [12].

The integers can be thought of as colors, and our objective in graph coloring is to make sure no two adjacent vertices of a graph are assigned the same color, given a maximum number of colors $q$. The minimum value of $q$ for a graph is known as the chromatic number of the graph. A graph is said to be $q$-colorable if $q$ colors can be assigned to the vertices without any clashes. A clash is when two connected vertices are assigned the same color.

### 2.2     Industry application

Graph coloring can be useful in the industry in several ways. One common application of graph coloring is in resource allocation using interval graphs [19]. The problem will start with resource requests.

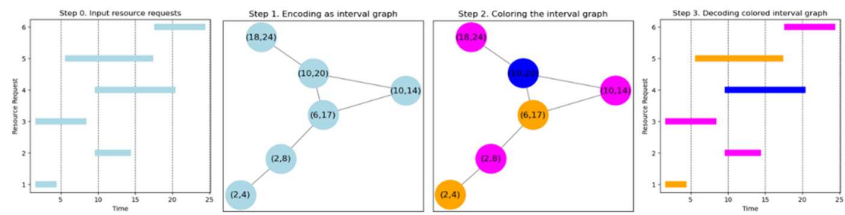

**Fig. 1.** Interval graph solution to resource allocation

Step 0 shows 6 resource requests. On x axis, we see the time duration of each resource request. Request 1 has a time duration from 2 to 4. request 2 from 10 to 14, request 3 from 2 to 8, request 4 from 10 to 20, request 5 from 6 to 17 and request 6 from 18 to 24.

Step 1 shows the resource request encoded as an interval graph. The six vertices represent the six requests. The labels of the vertices represent the time interval of each request. An edge between two vertices represents an overlap between the time interval of the two resource requests.

Step 2 shows the interval graph with colorings applied. Each color represents a resource. No two connected vertices have the same color, which implies that no two overlapping resource requests are assigned the same color.

In step 3, we decode the colored interval graph. Resource request 1 and 5 are assigned the same resource A (represented by orange) since they do not overlap, resource requests 2, 3, and 6 are assigned the same B (represented by magenta) since they do not overlap, resource request 4 is assigned resource C (represented by blue).

Similarly, problems such as scheduling can also be framed as graph coloring problems [16]

## 3    Heat Diffusion Framework

### 3.1    Preliminaries

The heat diffusion framework is a gradient based iterative solver. In the heat diffusion framework, each parameter is referred $\theta$ to as the location. Each location is associated with an initial temperature value $h(\theta)$. Instead of the solver having to look through a large space, the heat from all locations flows to the solver. This flow of heat allows the solver to find $\theta^*$ where the maxima of the temperature $h(\theta^*)$ lies. The details and proofs for this framework can be found in [13].

At any time $\tau$ and location $\theta$, the temperature distribution is given by $u(\tau, \theta)$. The gradient of the temperature is given by the following equation.

$$\nabla_\theta u(\tau, \theta) \approx \frac{1}{M} \sum_{m=1}^{M} \nabla_\theta f \left( erf \left( \frac{\theta - x^m}{\sqrt{2\tau}} \right) \right) \tag{1}$$

$erf(.)$ is the error function used to transform continuous $\theta$ to binary. M is the number of samples which is set to 1. $f$ is the target function that needs to be minimized. By creating the right target function $f$, the heat diffusion framework can be applied to a variety of problems. To run the gradient based iterative solver, $x_t$ can be sample from a uniform distribution over $[0,1]^n$. Gradient can be calculated using (1). Then θ can be updated by projecting it over a valid interval $[0,1]^n$. This process can be repeated for $T$ iterations.

### 3.2    Target function for graph coloring.

For graph coloring problem. Given an adjacency matrix $A$ for the graph. The target function can be defined as follows:

$$f(x) = sum\left(A * softmax\left(\frac{x}{\alpha}\right) * softmax\left(\frac{x}{\alpha}\right)^T\right) \quad (2)$$

Here, $x$ is initialized as a matrix of dimensions *(n,k)*. Where *n* is the number of vertices in the graph and *k* is the chromatic number of the graph. The final value of $x$ is decoded by taking the argmax over rows. The argmax indicates the color of the vertex.

## 4 Experiments

We obtained 33 graphs from [20]. The chromatic number for these graphs is known. We compare our methods against two other methods. First is greedy method. In the greedy method we parse the vertices of the graph with the largest-first strategy and greedily assign colors. If the chromatic number $k$ of the graph is reached, then we assign a dummy color. Hence, by design, the graph cannot have more than $k$ colors, with the exclusion of the dummy color. The details of implementation of greedy method can be found in [21]. The second method we compare our results against is TabuCol. TabuCol is a Tabu search based heuristic. The details for this method can be found in [22].

The performance of the methods was measured in terms of percentage of edges that are clashing. An edge is said to be clashing if it connects two vertices of the same color. Also, in case of the greedy method is said to be clashing if either of its vertices is assigned the dummy color. The dummy-colored vertices can be thought of as vertices to which the greedy method could not assign a color.

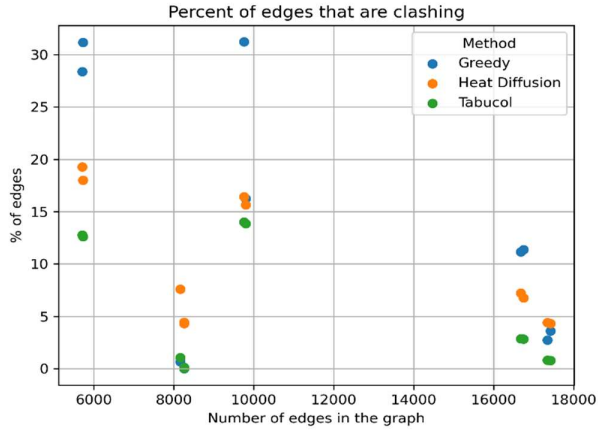

**Fig. 2.** Percent of edges that are clashing versus number of edges in the graph

We see that the percent of edges that are clashing is the lowest for TabuCol, followed by heat diffusion and greedy across different number of edges.

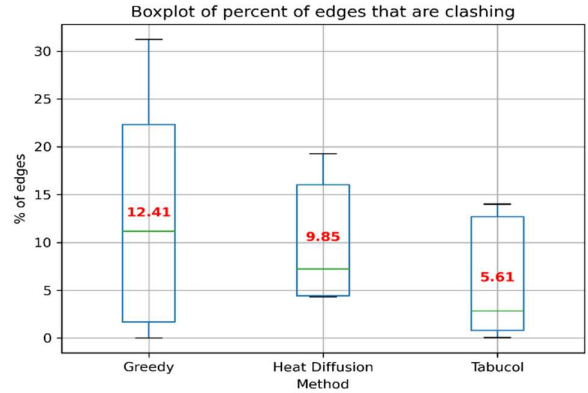

**Fig. 3.** Boxplot of percent of edges that are clashing with mean values

In figure 3 we see that the mean of percent of edges that are clashing is the lowest for TabuCol, second lowest is heat diffusion, and the highest is greedy method.

## 5   Conclusion

In this paper we put forth a new method for solving graph coloring using heat diffusion. By comparing the percent of clashes in the colored graph, we conclude that the heat diffusion method is competitive with popular methods.